\documentclass[10pt,twocolumn,letterpaper]{article}

\usepackage[pagenumbers]{wacv} % To force page numbers, e.g. for an arXiv version

\usepackage{graphicx}
\usepackage{amsmath}
\usepackage{amssymb}
\usepackage{booktabs}
\usepackage{svg}
\usepackage[accsupp]{axessibility}

\usepackage[pagebackref,breaklinks,colorlinks]{hyperref}

\usepackage[capitalize]{cleveref}
\crefname{section}{Sec.}{Secs.}
\Crefname{section}{Section}{Sections}
\Crefname{table}{Table}{Tables}
\crefname{table}{Tab.}{Tabs.}

\def\wacvPaperID{1194} % *** Enter the WACV Paper ID here
\def\confName{WACV}
\def\confYear{2025}

\begin{document}

\title{Joint Co-Speech Gesture and Expressive Talking Face Generation using Diffusion with Adapters}

% \and

\author{Steven Hogue \enskip Chenxu Zhang \enskip Yapeng Tian \enskip Xiaohu Guo\\
University of Texas at Dallas\\
{\tt\small \{ditzley, chenxu.zhang, yapeng.tian, xguo\}@utdallas.edu}
}

\maketitle

%% 8 page limit

%%%%%%%%% ABSTRACT
\begin{abstract}
Recent advances in co-speech gesture and talking head generation have been impressive, yet most methods focus on only one of the two tasks. Those that attempt to generate both often rely on separate models or network modules, increasing training complexity and ignoring the inherent relationship between face and body movements. To address the challenges, in this paper, we propose a novel model architecture that jointly generates face and body motions within a single network. This approach leverages shared weights between modalities, facilitated by adapters that enable adaptation to a common latent space. Our experiments demonstrate that the proposed framework not only maintains state-of-the-art co-speech gesture and talking head generation performance but also significantly reduces the number of parameters required.
\end{abstract}
\vspace{-5mm}
%%%%%%%%% BODY TEXT
\section{Introduction}
Co-speech gestures play a vital role in communication, adding additional context, emphasis, or emotion to the spoken words of a person. Similarly, the expressions of a person's face can add the same information, and in absence of these two things, listeners are left only relying on the words and tone of a person's speech. Audio-driven co-speech gesture generation and expressive talking heads are becoming significantly more interesting as the digital landscape expands to include applications such as AI chat-bots, virtual worlds, and online video conferencing. In these applications, being able to go from audio-only to realistic gesturing and speaking virtual avatars can improve usability and expressiveness. 

However, co-speech gestures and talking head are often problems that existing methods try to solve separately. Many methods either focus on co-speech gestures such as \cite{ginosar2019learning, zhu2023taming}, or just on the face such as \cite{Thambiraja_2023_ICCV}. Some methods model both at the same time but with completely different and separate architectures such as \cite{yi2022generating}, or slightly different but joint architectures such as \cite{chen2024diffsheg}. When only modeling only co-speech gestures or only talking face, the models potentially miss out on the weak correlation that gesture and facial motion have with each other. Of course, modelling only one also means that the other part needs to be either ignored or modelled separately somehow. Similarly, the methods that attempt to unify both in a single pipeline, but ultimately keep them separate suffer from the same issues that only modelling one has, they keep the body and face disjoint.

Keeping the two problems separate means having to utilize two networks. Both networks need to be trained, and both networks need to be used at inference time, potentially increasing hardware costs and development time. The naive solution to using two completely separate networks is to use two connected networks. However, this only solves the issue of leveraging what may be only a weak correlation between gesture and facial motion. This method still leaves increased parameter count and thus requires larger amounts of memory. Instead what if a single network could be used? Training in a single network for two tasks seems doomed to fail. Recently, large foundational models such as LLMs \cite{houlsby2019parameterefficient, hu2022lora} or text-to-image diffusion models \cite{luo2023lcmlora} have been adapted to perform tasks on which they were not originally trained. Utilizing fine-tuned modules inserted into a pre-trained network, these methods are able to achieve good results with minimal retraining and only a small increase in parameter count. 

In this work, we propose to introduce the idea of adapter modules to not adapt a pre-trained model, but instead to allow the use of a single network for modeling two weakly correlated but otherwise independent tasks at the same time. In our work we have a single transformer-based denoising network with jointly trained adapter modules to model both co-speech gestures and an expressive talking face. Our method allows the the face and body to influence each other while reducing the amount of parameters that would be present in two separate or two connected networks. In this way, we can produce state-of-the-art results for both co-speech gestures and expressive talking face while also maintaining a smaller parameter footprint than using two separate networks.

Our contributions can be summarized as follows: 
\begin{itemize}
    \item We introduce adapter modules to combine both the co-speech gestures and expressive talking face into a single network, significantly reducing the number of required parameters to effectively model both tasks in two separate networks.
    \item Our adapter modules also allow for the gestures and facial motions to influence each other, leveraging the weak correlation that they share.
    \item Our diffusion-based method is able to achieve state-of-the-art results for both co-speech gestures and expressive talking face both quantitatively and qualitatively, with a user study validating our method as having higher performance than existing methods in terms of realism and believability.  
\end{itemize}

The code is available at \url{https://github.com/Ditzley/joint-gestures-and-face}.

\vspace{-4mm}
\section{Related Works}
\subsection{Co-Speech Gestures}
Work focusing on co-speech gestures have used many different input and output modalities and network architectures. Recent methods have shifted focus to using data-driven methods in order to better reconstruct the speech-to-gesture mapping. This many-to-many mapping has encouraged the use of generative models, including the use of transformers \cite{sun2023co, pang2023bodyformer}, recurrent neural networks (RNNs) \cite{yoon2019robots, ao2022rhythmic}, generative adversarial networks (GANs) \cite{ginosar2019learning,yoon2020speech,liu2022learning}, variational auto-encoders (VAEs), \cite{li2021audio2gestures, liu2022audio}, flow-based models \cite{alexanderson2020style,ye2022audio}, and most recently diffusion based models \cite{zhu2023taming,ao2023gesturediffuclip,yang2023DiffuseStyleGesture, Deichler_2023}. The diffusion based methods specifically leverage the powerful diffusion models to introduce diversity into the generated gestures while also maintaining the high quality output that is competitive with other state-of-the-art methods. However, these methods all focus on the generation of only co-speech gestures, which leaves out the an important part of human speech, the face.

\subsection{Talking Head}
Talking head generation lends itself to more traditional models for reconstruction. While co-speech gestures have a more obvious many-to-many mapping from speech to gesture, talking head is a bit more rigid. Many previous methods have focused on image generation \cite{zhang2022sadtalker, shen2023difftalk}. Some methods have chosen to instead animate a 3D face by directly deforming face vertices or learning face motion parameters \cite{faceformer2022, xing2023codetalker, Thambiraja_2023_ICCV}. The focus on talking head, similarly to the co-speech gesture works, leaves out a significant portion of the talking human system, the body.

\subsection{Combined Co-Speech Gestures and Talking Head}
Combining both the body and face and generating both co-speech gestures and a talking head is not a trivial task. A few works have generated both, such as \cite{liao2020speech2video, qian2021speech}. More recent and sophisticated methods have also tried to tackle generation of both gesture and talking head, such as TalkSHOW \cite{yi2022generating}, but use separate models to generate the two sequences and combine them later. The disjoint structure of TalkSHOW requires the implementation and training of separate face (VAE) and body models (VQ-VAE), with no connection between the two except for the audio input. In an attempt to let the parts influence each other, another work EMAGE \cite{liu2024emage}, jointly trains separate VQ-VAE models for upper body gestures, hands, lower body, and face, with some connection through cross-attentions. Similarly, DiffSHEG \cite{chen2024diffsheg}, trains a diffusion model to generate both talking head and co-speech gestures, using the talking head as further conditioning on the gesture generations. These three works, however, all share the same downside of using separate networks for talking head and body gestures. In our work, we combine the talking head and co-speech gestures into a single network with shared weights between both.

\section{Method}
\begin{figure*}
    \centering
    \includegraphics[width=\linewidth,trim={0 1.25cm 0 2.5mm}, trim]{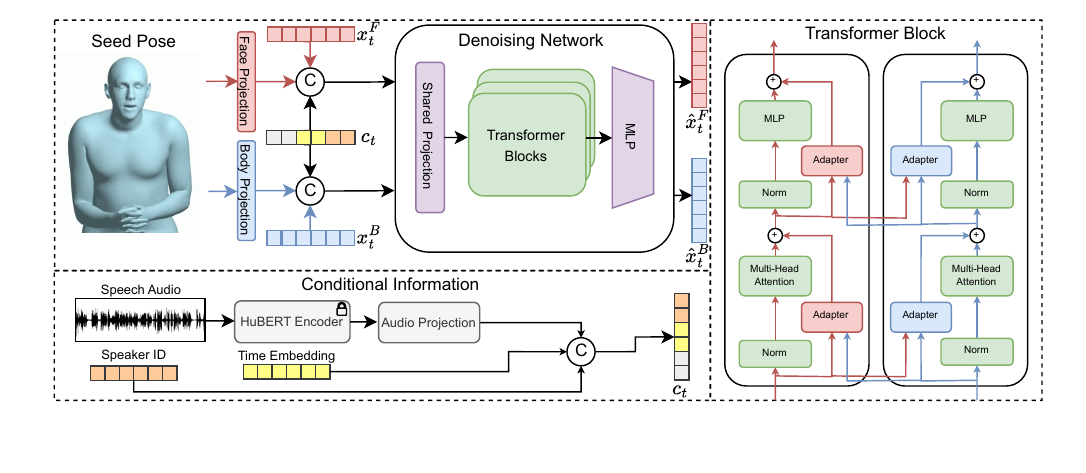}
    \caption{\textbf{Framework and Architecture Overview.} The \textbf{top left} shows the general network architecture, with inputs of a face motion parameters and body parameters being fed into separate projection layers, combined with conditional information $c_t$ from the \textbf{bottom left}, and the noisy latents $x_t$. The transformer network (in green) denoises the latents, which are fed into the network separately. The \textbf{right} is the architecture of our transformer block with adapters. The green transformer blocks share one set of parameters for both branches.}
    \label{fig:pipeline}
    \vspace{-5mm}
\end{figure*}
Modelling the talking head and co-speech gestures jointly can be done naively by simply concatenating the two together and training a diffusion model to generate this combined sequence. While this method has the advantage of reducing the number of networks or separate modules needed for generating a sequence, thus reducing total parameter count but ultimately leads to poorer results as compared to separate networks. We, instead, propose a method of using a single diffusion network to diffuse both talking head and co-speech gestures from a shared latent space. Our work utilizes adapter modules to enable information sharing between the two different modalities and help the main shared transformer network model both talking head and co-speech gestures jointly. 

\subsection{Problem Formulation}
Given arbitrary speech audio, we aim to generate diverse and realistic gestures with expressive talking face. Training samples from aligned audio and motion sequences are obtained using a sliding window of N frames, with M frame overlap. For each N-frame long clip, audio is encoded into audio features $\mathbf{A}=[a_1,a_2,...,a_N]$. The motion is split into face and body motions. The body motion, $\mathbf{B}=[b_1,b_2,...,b_N]$, is represented by joint-rotations, where $b_i \in \mathbb{R}^{3\times J}$ and $J$ is the number of joints. The face motion, $\mathbf{F}=[f_1, f_2, ..., f_N]$, is represented by the jaw joint, as joint rotation, and facial expressions, as blend shape weights, where $f_i \in \mathbb{R}^{3 + E}$ and $E$ is the number of expression parameters. The entire motion sequence is the concatenation of body motion, $\mathbf{B}$, and face motion, $\mathbf{F}$. The objective of our method is to reconstruct this motion conditioned on the audio, $\mathbf{A}$.

\subsection{Preliminaries}
To accomplish the this task, we introduce a diffusion model, following ~\cite{ho2020denoising}, augmented by an adapter module. We briefly introduce the concepts of both here. 

\vspace{-3mm}
\subsubsection{Diffusion}
To reconstruct the motion sequence conditioned on an arbitrary model, we implement a diffusion model. Diffusion works to model the distribution of the data, utilizing both the diffusion process and the denoising process. The goal of diffusion is when given some sample of the data, $\mathbf{x}_0$, from the real data distribution $q(\mathbf{x}_0)$, to learn a model distribution, $p_\theta(\mathbf{x}_0)$, that approximates the real distribution, $p(\mathbf{x}_0$, where, $\mathbf{x}_0$ represents the motion sequence. 

\vspace*{-\baselineskip}
\paragraph{Diffusion Process}
In the diffusion process we progressively add noise to the input data sample, $x_0\sim p(\mathbf{x}_0)$, based on a predefined variance schedule $\beta_1,...,\beta_T$, where $\beta_t\in (0, 1)$, and timestep $t={1,...,T}$. This iterative process can be defined as:
\begin{equation}
q(\mathbf{x}_t|\mathbf{x}_{t-1})=\mathcal{N}(\mathbf{x}_t;\sqrt{1-\beta_t}\mathbf{x}_{t-1},\beta_t\mathbf{I}).
\label{eq:q}
\end{equation}

\vspace*{-\baselineskip}
\paragraph{Denoising Process}
In the denoising process, we work in the inverse direction. The model is trained to invert the addition of the noise at each timestep, $t$, and iteratively remove the noise to reach a sample in the original data distribution. The denoising process can be defined as:
\begin{equation}
p_\theta(\mathbf{x}_{t-1}|\mathbf{x}_t,c)=\mathcal{N}(\mathbf{x}_{t-1};\mu_\theta(\mathbf{x}_t, t, \mathbf{c}), \beta_t\mathbf{I}),
\label{eq:ptheta}
\end{equation}
In this process, we must also inject the conditional information, $c$, so that the reconstruction of the data sample is conditional on this input information, rather than on strictly noise. This conditional information, which we discuss in more detail later, is the audio, $A$, an $M$-frame long seed gesture, and the speaker ID. 

\vspace*{-\baselineskip}
\subsubsection{Adapters}
Adapters have become relatively popular in natural language processing (NLP) tasks, specifically in large language models (LLMs) \cite{houlsby2019parameterefficient, hu2022lora}.
Additionally, diffusion models focusing on text-to-image (T2I) generation have begun to use them as well \cite{luo2023lcmlora}.
These adapters, as the name suggests, aim to adapt large pre-trained networks to perform additional tasks. Because, LLMs and T2I models are normally trained on very large datasets and often have billions of parameters, retraining the models can be computationally expensive and time consuming \cite{liu2019roberta, brown2020language, he2021debertav3, podell2023sdxl}. 
Adapters are small modules that are then inserted, after training, into the larger networks. Fine-tuning of these modules, while the main network's parameters are frozen, is much less expensive in terms of computation and time, than retraining a new model for each task. 

The architecture of adapter modules are typically bottleneck networks with only a few layers. Bottleneck networks allow a trade-off between parameter count and performance, as discussed in \cite{houlsby2019parameterefficient}.  
The placement of such adapter modules are also a topic of research, with some papers putting them between layers of the transformer blocks \cite{houlsby2019parameterefficient, Rebuffi17, lin2020exploring}
while others, such as LoRA \cite{hu2022lora}, place the adapters in parallel to reduce latency during inference, as adapters between layers, are executed sequentially with the rest of the model.

These adapters have shown significant ability for enabling a network to transfer knowledge from one domain to another, with only the need for quick fine-tuning of a small number of parameters. This allows the reuse of large pre-trained networks to tackle other domains. In our work, we explore the use of an adapter module without a pre-trained network to, instead, help map between two related but fundamentally different tasks, co-speech gesture and talking face generation. 

\subsection{Framework}
Our framework consists of a main diffusion model and the inserted adapter modules. The main diffusion model is trained to reconstruct a motion sequence, including both the body gestures and the face movement and expressions. The diffusion model aims to reconstruct this motion sequence in a parameter-efficient way by utilizing the adapter modules. 

\noindent
\textbf{Conditional Information.}
\label{sec:cond}
The diffusion model is further conditioned on extra information, an initial $M$-frame long seed gesture, a speech audio, and additionally a speaker ID. The seed gesture's primary function is to assist the model in setting the initial pose of the motion sequence. It also has the added benefit of allowing the connection of multiple segments of generated gestures, which we discuss later. 

Rather than using the raw audio as a condition for the model, we instead convert it into high-level speech features. The audio is encoded using the HuBERT speech encoder \cite{hsu2021hubert}. HuBERT focuses on speech-to-text tasks and thus is able to capture high-level semantic information that is fundamental to producing believable face movements. During training, the HuBERT encoder is frozen and instead a single linear layer is trained to project from the encoder's larger feature dimension to a smaller dimension to be fed into the transformer network. All conditional information, the seed gesture, audio features, and speaker ID, are concatenated together with the timestep embedding and injected into the transformer network.

\noindent
\textbf{Shared Transformer Network.}
\label{sec:transformer}
The transformer network works to denoise the motion sequence segment given to it, with guidance of the conditional information. 
The network consists of a transformer encoder block and a transformer decoder block. Both the transformer encoder and decoder are created as a stack of transformers, and are connected together through a LayerNorm. Additionally, a LayerNorm is placed after the transformer decoder block and feeds into a final MLP. 

Rather than concatenating the face motion and body gestures together and feeding them into the network, we approach them separately, but in the same way. The face and body each have individual linear layers to project them into a shared latent space. These are then fed into another linear layer to a final shared latent dimension. The face and body features are then separately concatenated with the conditional information and fed into the transformer network in parallel. The transformer network then outputs a separate face and body latent sequences. An additional linear layer for each the face and body then project them separately back into the body gesture and facial motion spaces. 

\noindent
\textbf{Adapter Module.}
Our adapter module works to help translate the face and body information to the shared latent space. We follow the work of LAVisH \cite{LAVISH_CVPR2023} to create our adapter. The adapter module is constructed as two cross-modal attention layers in sequence that feed into a bottleneck that consists of a downsample layer, an activation layer, a hidden layer, and finally an upsample layer. There are four adapter layers per transformer block layer, in both the encoder and decoder. Two of the layers are for the face motion and two are for the body gesture. 

The face and body sequences being completely separate discounts the ability for them to influence and interact with each other. The production of facial motion and body gestures are done in tandem rather than in isolation. As such, the cross-modal attention layers are added to facilitate this cooperation. In the first attention layer, the other sequence is combined with trained latent tokens, that is for the face adapter the body sequence is used as input, and vice versa. The output of this layer is then combined with the relevant sequence in the second attention layer. In this way, we can fuse the features from both face and body together in each side of the network, so that they can influence each other. The fused features are then fed into the remaining layers of the adapter module and then combined with the transformer network. 

\subsection{Training}
\paragraph{Loss Functions}
Diffusion models are typically trained, following \cite{ho2020denoising}, by optimizing the variational lower bound on negative log-likelihood: $\mathbb{E}[-\log p_\theta(\mathbf{x}_0)]\leq\mathbb{E}_q[-\log\frac{p_\theta(\mathbf{x}_0)}{q(\mathbf{x}_{1:T}|\mathbf{x}_0)}]$. This can be simplified to the loss function:
\begin{equation}
    L=\mathbb{E}[\|\mathbf{\epsilon}-\mathbf{\epsilon}_\theta(\mathbf{x}_t,\mathbf{c},t)\|^2],
    \label{eq:noise-loss}
\end{equation}
where $\epsilon\sim\mathcal{N}(0, \mathbf{I})$ is Gaussian noise and $\epsilon_\theta(\mathbf{x}_t,\mathbf{c},t)$ is the network predicting the noise to remove. Then the resulting prediction at timestep, $t$, can be defined as $\mathbf{x}_t=\sqrt{\bar{\alpha}_t}\mathbf{x}_0+\sqrt{1-\bar{\alpha}_t}\mathbf{\epsilon}_t$, where  $\alpha_t=1-\beta_t$ and $\bar{\alpha}_t=\prod_{s=1}^{t}\alpha_s$. 

Instead of predicting the noise, we follow \cite{ramesh2022,tevet2023human,tseng2023edge} and directly predict the sample at each timestep. This approach improves the stability of the final prediction and allows for the seamless addition of an extra loss function.Predicting the noise would require calculating the original sample $x_0$ from $x_t$ before applying the loss function, which could introduce additional error. By directly predicting $x_0$ at each timestep, we avoid this potential error and iteratively refine the predicted sequence.

Doing this, we can replace the noise in \cref{eq:noise-loss} with the sequences, $\mathbf{x_0}$ and 
$\hat{\mathbf{x}}_0=\mathbf{x}_\theta(\mathbf{x}_t,\mathbf{c},t)$, 
 directly to give us our first loss function, the reconstruction loss:
\vspace{-2mm}
\begin{equation}
L_{rec}=\mathbb{E}[\|\mathbf{x}_0-\hat{\mathbf{x}_0}\|^2].
\label{eq:rec-loss}
\vspace{-2mm}
\end{equation}

We then add a velocity loss which is calculated as the mean absolute error of the velocity between frames: 
\begin{equation}
L_{vel}=\mathbb{E}[\|(\mathbf{x}_0[1:]-\mathbf{x}_0[:-1])-(\hat{\mathbf{x}}_0[1:]-\hat{\mathbf{x}}_0[:-1])\|].
\label{eq:vel-loss}
\end{equation}
The velocity loss serves as a smoothing term to help the network produce temporally consistent predictions and ultimately, smooth motions between adjacent frames. 

\subsection{Inference}
During training, we segmented audio and motion videos into clips of a fixed length $N$, for inference, this fixed length is not ideal, as we want to be able to produce gesture sequences of much longer lengths to match the input audios. To do this we can leverage the $M$-frame long seed gesture used as conditional information. For long clips, we split audios with a sliding window into $N$-frame long segments with $M$-frame overlap, much like in the preparation of the training data. However, in contrast to the training data, at inference we take the seed gesture for only the first segment. We then perform inference on the first segment, that includes the seed gesture, and for the remaining segments we take the last $M$-frames of the previous segment as our seed gesture. In this way we can produce motion sequences much longer than the $N$-frames used for training.

However, a problem arises, the diffusion model is unable to exactly replicate the seed gesture as the first $M$-frames. This is in part a strength of the diffusion model to increase the diversity of the results by including additional noise at inference time in addition to starting the denoising process from pure noise. This leads to a disconnect between segments that can be alleviated by interpolation of the final $M$-frames and first $M$-frames of adjacent segments: 
\begin{equation}
\mathbf{x}_i=\mathbf{x}_{prev,i} * \frac{M - i}{M + 1} + \mathbf{x}_{next,i} * \frac{i + 1}{M + 1},
\label{eq:interp}
\end{equation}
where $\mathbf{x}_{prev,i}$ and $\mathbf{x}_{next,i}$ are the $i$th frame of the overlap between the first and second sequences respectively, $i\in\{0,...,M-1\}$, and $x_i$ is part of our final output sequence.

\section{Experiments}
\begin{figure*}[ht]
    \captionsetup[subfigure]{labelformat=empty}
    \centering
    \begin{subfigure}{0.33\linewidth}
        \includegraphics[width=\linewidth,trim={0 5mm 0 10.55cm},clip]{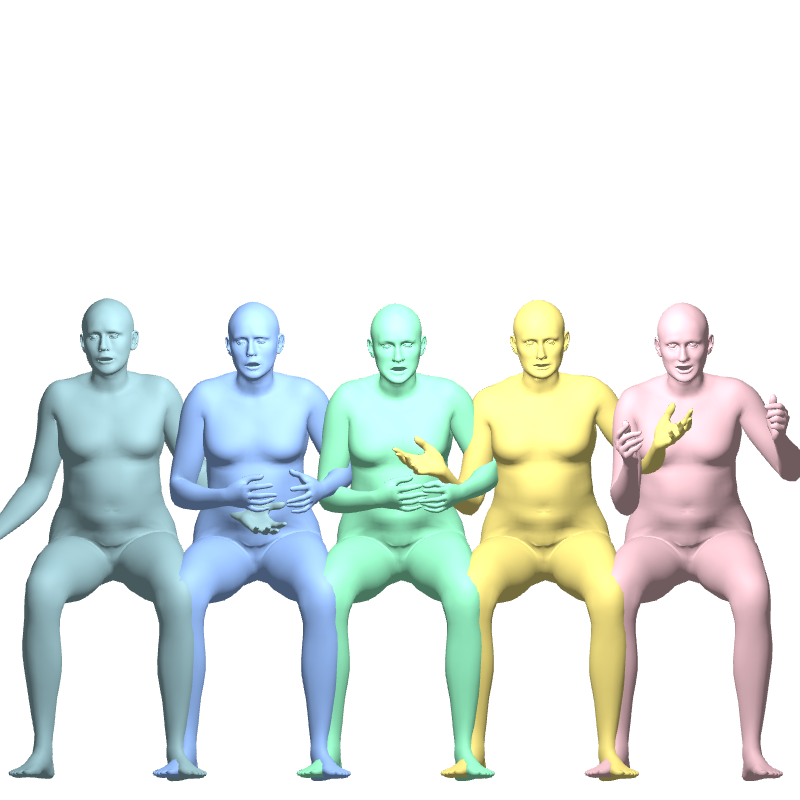}
        \caption{Ours}
    \end{subfigure}%
    \begin{subfigure}{0.33\linewidth}
        \includegraphics[width=\linewidth,trim={0 5mm 0 10.55cm},clip]{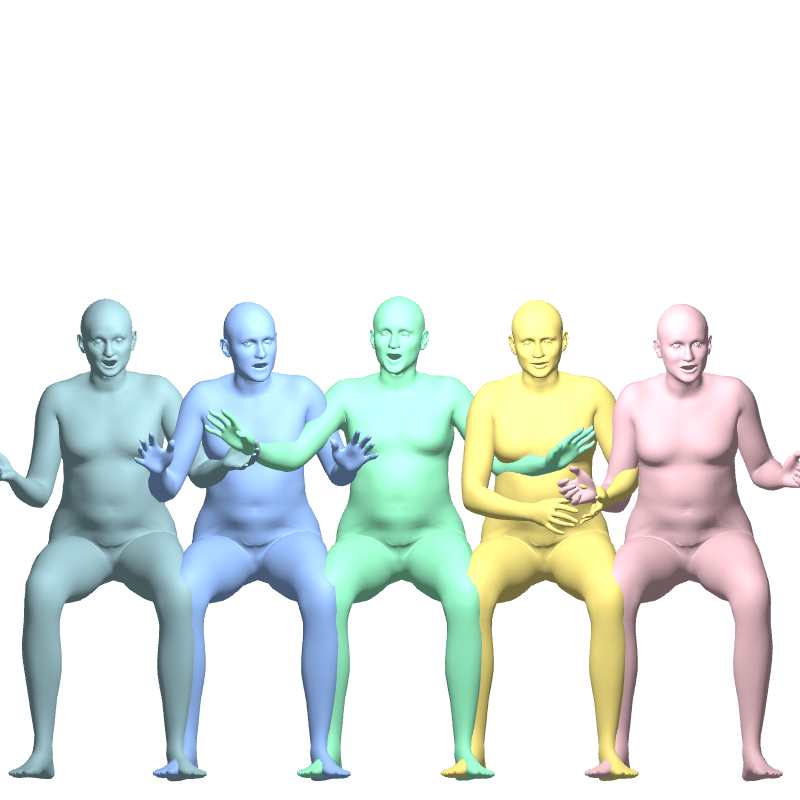}
        \caption{TalkSHOW}
    \end{subfigure}
    \begin{subfigure}{0.33\linewidth}
        \includegraphics[width=\linewidth,trim={0 5mm 0 10.55cm},clip]{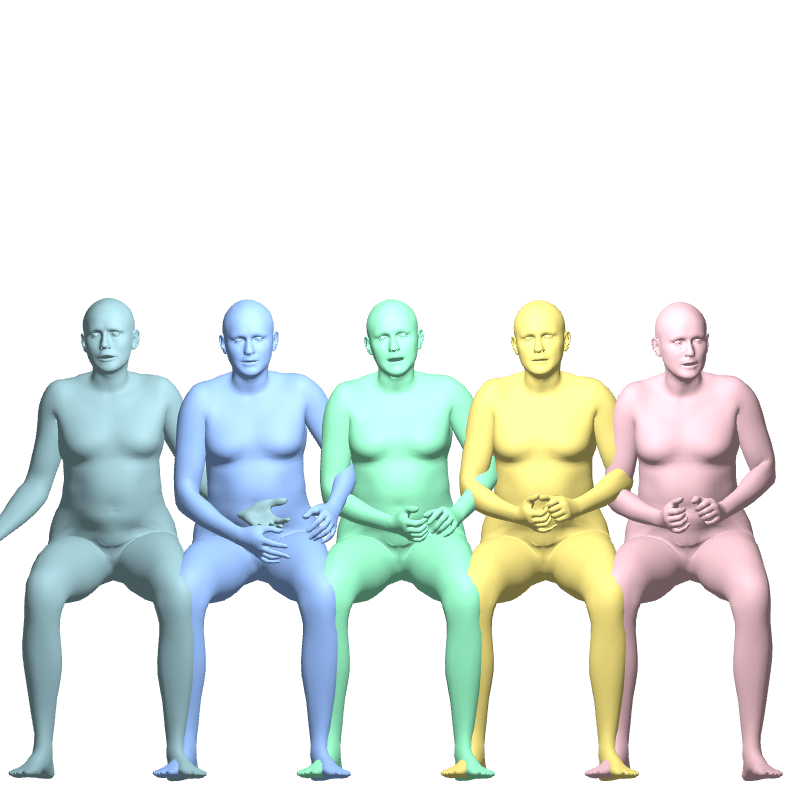}
        \caption{DiffGesture}
    \end{subfigure}
    \caption{\textbf{Qualitative Comparison.} We compare sequences of motions for our method, TalkSHOW and DiffGesture. Our motions are more diverse and dynamic compared to the baselines.}
    \label{fig:qual}
\end{figure*}

\label{sec:quant}
\begin{table*}[t]
    \centering
    \begin{tabular}{l | c c c | c c c c c | c}
        \toprule
        Methods & FMD$\downarrow$ & Div (All)$\uparrow$ & BC$\uparrow$ & FED$\downarrow$ & Div (Face)$\uparrow$ & Jaw L1$\downarrow$ & Lmk L1$\downarrow$ & LVD$\downarrow$ & Params \\
        \midrule
        LS3DCG \cite{ls3dcg} & 2185.23 & 1314.38 & \textbf{0.791} & 1618.83 & 1064.19 & \textbf{0.00147} & \textbf{0.1410} & \textbf{0.0273} & 18.7M \\
        DiffGesture \cite{zhu2023taming} & 5831.43 & 1405.14 & 0.758 & 5000.59 & 1176.17 & 0.00192 & 0.2088 & 0.0333 & 51.6M \\
        TalkSHOW \cite{yi2022generating} & 3681.39 & 1704.25 & 0.733 & 3019.28 & 1387.04 & 0.00191 & 0.2174 & 0.0294 & 120M \\
        DiffSHEG \cite{chen2024diffsheg} & 1882.10 & \textbf{1924.78} & 0.755 & 1379.55 & \textbf{1609.03} & 0.00189 & 0.1821 & 0.0298 & 155M \\
        Ours & \textbf{1758.13} & 1845.15 & 0.763 & \textbf{1260.01} & 1521.68 & 0.00161 & 0.1532 & 0.0276 & 27.7M \\
        \bottomrule
    \end{tabular}
    \caption{\textbf{Quantitative Results.} We compare between our method and a GAN method LS3DCG, a diffusion-based method DiffGesture, a VQ-VAE method TalkSHOW, and a diffusion-based method DiffSHEG.}
    \label{tab:quant}
    \vspace{-5mm}
\end{table*}

\begin{figure*}
    \captionsetup[subfigure]{labelformat=empty}
    \centering
    \begin{subfigure}{0.33\linewidth}
        \includegraphics[width=\linewidth,trim={0 5mm 0 10.55cm},clip]{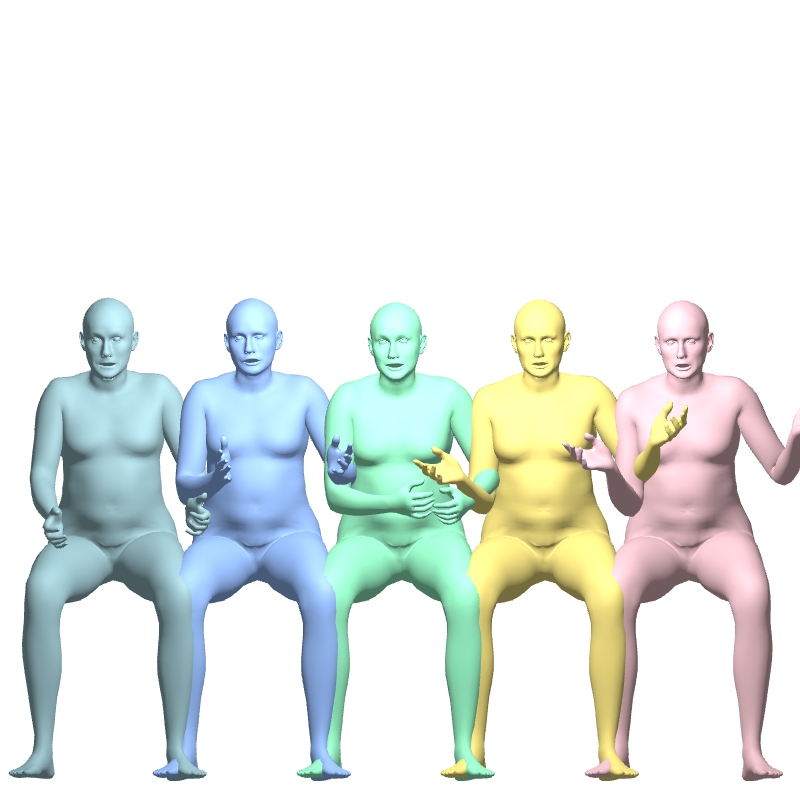}
        \caption{Ours}
    \end{subfigure}%
    \begin{subfigure}{0.33\linewidth}
        \includegraphics[width=\linewidth,trim={0 5mm 0 10.55cm},clip]{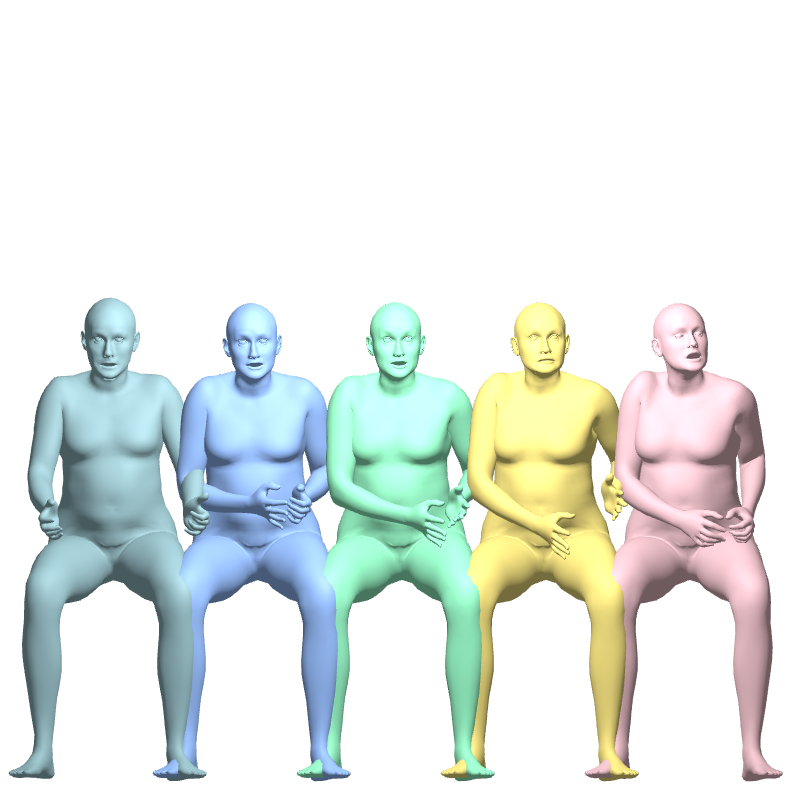}
        \caption{Separate}
    \end{subfigure}%
    \begin{subfigure}{0.33\linewidth}
        \includegraphics[width=\linewidth,trim={0 5mm 0 10.55cm},clip]{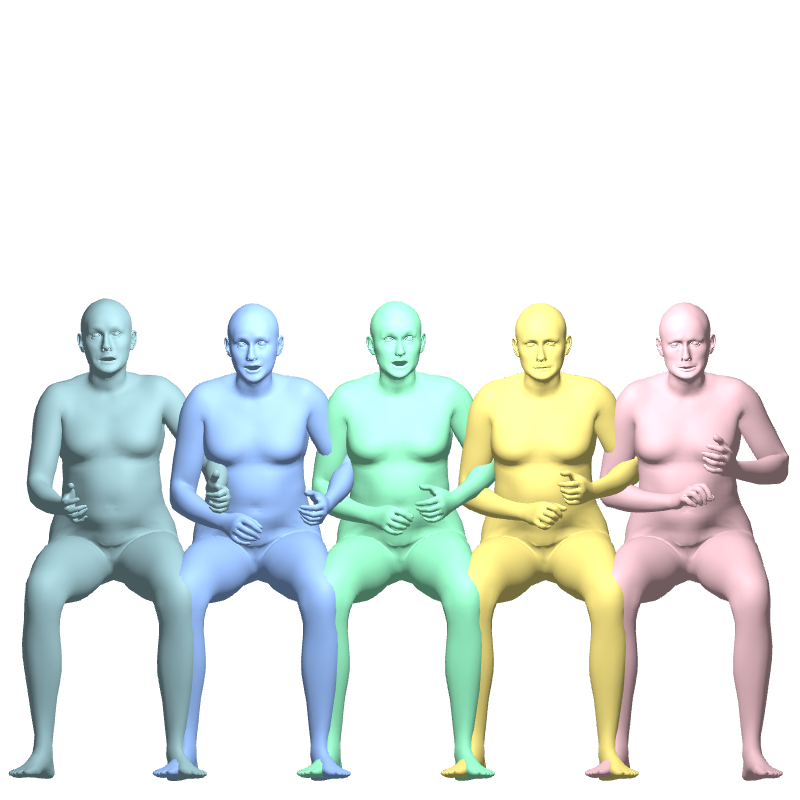}
        \caption{Combined}
    \end{subfigure}
    
    \vspace{-3mm}
    \caption{\textbf{Qualitative Comparison for Ablation Study.} Our model produces smooth dynamic motion while the alternative architectures generate jittery motions that move little from the mean position.}
    \label{fig:ablat}
\end{figure*}

\begin{table*}[t]
    \centering
    \begin{tabular}{l | c c c | c c c c c | c}
        \toprule
        Methods & FMD$\downarrow$ & Div (All)$\uparrow$ & BC$\uparrow$ & FED$\downarrow$ & Div (Face)$\uparrow$ & Jaw L1$\downarrow$ & Lmk L1$\downarrow$ & LVD$\downarrow$ & Params \\
        \midrule
        Separate & \textbf{1525.92} & \textbf{2043.56} & 0.764 & 1235.55 & \textbf{1698.91} & 0.00174 & 0.1671 & 0.0289 & 51.2M\\
        Combined & 1560.36 & 1936.99 & \textbf{0.772} & \textbf{994.21} & 1580.54 & 0.00172 & 0.1641 & 0.0278 & 25.6M \\
        Split & 9420.91 & 1189.83 & 0.747 & 8450.66 & 866.59 & 0.00212 & 0.2224 & 0.0387 & 53.1M \\ 
        Ours & 1758.13 & 1845.15 & 0.763 & 1260.01 & 1521.68 & \textbf{0.00161} & \textbf{0.1532} & \textbf{0.0276} & 27.7M \\
        \bottomrule
    \end{tabular}
    \caption{\textbf{Ablation Study.} We compare against two setups that do not include the adapter module, a combined model, where face and body are predicted together, and a separate model, where the face and body are predicted separately. We also compare against a split model, where two transformer models are connected by the adapter module.}
    \label{tab:ablat}
    \vspace{-5mm}
\end{table*}

\begin{table*}[t]
    \centering
    \begin{tabular}{l | c c c c c}
        \toprule
        Methods & Gesture Realism & Gesture Smoothness & Gesture Sync & Lip Sync & Expression Realism \\
        \midrule
        GT & 4.63 & 4.50 & 4.88 & 5.00 & 4.88 \\
        \midrule
        LS3DCG \cite{ls3dcg} & 2.50 & 1.75 & 2.38 & 2.63 & 2.63 \\
        DiffGesture \cite{zhu2023taming} & 2.00 & 1.50 & 1.50 & 1.38 & 1.38 \\
        TalkSHOW \cite{yi2022generating} & 2.88 & 3.00 & 2.75 & 3.00 & 3.00 \\
        DiffSHEG \cite{chen2024diffsheg} & 2.88 & 3.25 & 3.25 & 3.5 & 3.63 \\
        \midrule
        Separate & 3.00 & 2.88 & 3.50 & 3.38 & 3.25 \\
        Combined & 2.50 & 2.00 & 2.63 & 2.75 & 2.38 \\
        Split & 2.13 & 2.38 & 2.00 & 2.00 & 2.00 \\
        \midrule
        Ours & \textbf{3.50} & \textbf{3.63} & \textbf{3.88} & \textbf{4.13} & \textbf{4.00} \\
        \bottomrule
    \end{tabular}
    \caption{\textbf{User Study.} In our user study we ask users to grade on several criteria to measure the realism and believability of the generated co-speech gestures and facial motions. We compare our method to both our baselines and our alternative models from the ablation studies, as well as to the ground truth data.}
    \label{tab:user}
    \vspace{-5mm}
\end{table*}

\subsection{Dataset}
Our model is trained on the SHOW dataset, which includes the body gestures for the hands and upper body, and facial motion consisting of the jaw motion and facial expression. For the body there are $J=43$ number of joints ($13$ for the body and $30$ for the hands) and for the face there is one jaw joint and $E=100$ blend shape weights for the expression. In total that gives us $(J + 1)\times 3 + E=232$ total parameters for the network to predict. We split sequences into $N=34$ frame clips with an $M=4$ frame overlap for the seed gesture.
The dataset splits are done following \cite{yi2022generating}, the SHOW dataset is split into 80\%/10\%/10\% for train/val/test sets, and only using videos that are longer than 3s.

\subsection{Qualitative Results}
As shown in \cref{fig:qual} we can see that our method creates a wide range of motions throughout a sequence compared to TalkSHOW and DiffGesture. DiffGesture often starts with decent gestures due to taking a seed pose as input, however, throughout a sequence it tends towards the mean pose position, generating dull poses with little diversity. TalkSHOW does not take a seed pose but still fails to create diverse motion in a single sequence, instead moving little from the starting position throughout a sequence. LS3DCG also does not take a seed pose as input but typically tends towards the mean pose as in the DiffGesture example. Additionally, the LS3DCG and DiffGesture motions suffer from jittery and choppy movements. TalkSHOW, while able to produce smooth gestures and face motions, moves little from the starting position.

The baseline methods produce dull motions with little variation from either mean pose or starting pose of the sequence and often suffer from jittery motions, while our method is able to produce dynamic and smooth motions that have greater diversity throughout the entire sequence.

\subsection{Quantitative Results}
\noindent
\textbf{Evaluation Metrics}
Our model aims to believable body gestures but not necessarily match the ground truth exactly, because body gestures are only weakly correlated to the speech audio. To this end, the body metrics aim to measure the realism and diversity of the generated gesture in comparison with the real gestures. In contrast to the body gesture, the face should match closely to the ground truth, as the motion of the face is highly correlated to the speech audio. Therefore, with the face our metrics aim to measure the similarity to the ground truth as well as to measure generated diversity in the expressions.

\noindent
\textbf{Fr\'echet Motion Distance (FMD)} 
Based on Fr\'echet Gesture Distance (FGD) introduced in \cite{yoon2020speech}, where they define the FGD by training an auto-encoder on the gesture sequence to extract features of real gesture sequences. Generated sequences are then compared to the real sequences in the same way as in the Fr\'echet Image Distance \cite{NIPS2017_8a1d6947} and Fr\'echet Video Distance \cite{unterthiner2019fvd} methods. The main goal of this metric is to determine whether the generated motion is of high quality in comparison to the real data distribution. The Fr\'echet Motion Distance (FMD) was introduced in \cite{chen2024diffsheg}, where they define FMD as the measurement of the holistic motion, rather than on just the gesture. We choose to use FMD to measure our generated motion to capture the combination of and weak correlation between the generated face motion and body gestures. 

\noindent
\textbf{Fr\'echet Expression Distance (FED)} 
Also based on the Fr\'echet Gesture Distance, and introduced in \cite{chen2024diffsheg}, the FED measures the Fr\'echet distance on only the facial expressions. We choose to include this metric as further comparison on the face motions because the baseline methods include separated face architectures rather than a joint architecture as in our method.

\noindent
\textbf{Diversity (Div)}
The measure of diversity uses the same auto-encoders used in the FMD and FED metrics for the holistic motion and the face motion, respectively. This metric measures the range of motion that the generative model can produce and is calculated following \cite{yi2022generating}. Diversity is an important metric for generative models because methods generating gestures that tend majorly toward the mean can still score well in other metrics.

\noindent
\textbf{Beat Consistency (BC)}
For co-speech gestures, the alignment of gestures to the beat is important so as to match the cadence of gesture and speech together. This metric measures the correlation between motions and audio beat. We follow \cite{chen2019hierarchical} for measuring beat consistency by calculating the mean absolute angle change between joints in adjacent frames to get kinematic beats. Audio beats are then detected by onset strength and compared to kinematic beats to get the consistency score. In this metric, perfect beat consistency is not necessarily ideal and should instead also be compared to the beat consistency of the real data.  

\noindent
\textbf{Jaw L1}
For the speech motion, the jaw must match closely to be believable. The Jaw L1 metric measures the L1 distance for only the jaw joint between the ground truth motion and the generated motion. This partially measures lip sync between the generated motion and the speech audio.

\noindent
\textbf{Landmark L1 (Lmk L1)}
Since the focus of this work is not on creating a variety of expressions and emotions for the speech, but rather capturing the expression of the speaker, the Landmark L1 measures L1 distance between the facial landmarks. This also helps capture the accuracy of landmarks around the mouth, which in combination with the Jaw L1 metric, can describe the lip sync between the generated motion and the speech audio.

\noindent
\textbf{Landmark Velocity Distance (LVD)}
The Jaw L1 and Landmark L1 individually measure the accuracy of the jaw and other facial landmarks in comparison to the ground truth. These metrics measure frame-by-frame and thus do not give a measure of the temporal consistency of the generated motion. LVD combines both jaw and other facial landmarks, and calculates the velocity of each parameter between frames to compare. With this metric we can discern how smooth and temporally consistent the generated motion is.

\noindent
\textbf{Comparison with Baselines.} The results shown in \cref{tab:quant} show that our method can achieve state-of-the-art results against a variety of different baselines. The baselines include LS3DCG \cite{ls3dcg}, a GAN-based method, DiffGesture \cite{zhu2023taming}, a diffusion-based (DDPM) method, TalkSHOW \cite{yi2022generating}, a VQ-VAE-based method, and DiffSHEG \cite{chen2024diffsheg}, a diffusion-based (DDIM) method. Since DiffGesture only focuses on the upper body gestures we additionally train a separate model for the face. 
For the diversity metric, we randomly choose two sets of 500 reconstructed sequences to compare, following \cite{zhu2023taming}. For the FMD and FED metrics we train two autoencoders following \cite{liu2022beat}. 
In the FMD, FED, and diversity measures, our method outperforms all baselines (except DiffSHEG in diversity) showcasing that it creates diverse gestures of high quality. Diversity and quality are often correlated with a trade-off between the two. Methods can easily achieve high diversity by generating wildly unrealistic motions or on the other hand, they can achieve high quality by generating motions that stay close to the mean pose and have little variation. Our method, however, achieve both higher quality, so motions that are realistic, and high diversity, or a wide variety of motions. It is also notable that diversity and beat consistency both can also be skewed by non-smooth, jittery motions. This is because the beat consistency does not penalize incorrect motion beats that are not the nearest in proximity to the audio beat, therefore the jittery motions often align well to the beat because there are a higher number of motion beats in a small time frame. 

We also compare the trainable parameter counts for each model. Our model shows significant reduction and performance gain over the TalkSHOW and DiffGesture models, and only having more parameters than the LS3DCG model. Performance wise, since we are using a diffusion model, the inference time is naturally slower than LS3DCG and TalkSHOW. In comparison with DiffGesture, our method has similar performance as the two network approach. However, this could be addressed by more careful parallelization of the face and body inputs in our network. DiffSHEG uses 25-step DDIM diffusion as opposed to a 500-step DDPM and thus naturally has significant performance gains, though not nearly as performant as LS3DCG and TalkSHOW.

\subsection{Ablation Studies}
\label{sec:ablation}
In the ablations, we focus on two alternative architectures; combined face and body model, where the face and body parameters are concatenated together and run through a single network, and then separate face and body networks where the face and body are trained separately with same architecture. In both models, the network architecture uses the same architecture as our proposed method with the exclusion of the adapter modules.

Quantitative metrics for the ablations can be seen in \cref{tab:ablat}. Notably, our model performs worse in FMD, FED, beat consistency, and diversity than the two alternative models. However, in face metrics our model performs better. This performance gap illustrates how misleading metrics can be in terms of judging model performance. Both models result in jittery and/or dull motions for body gestures and facial movements. Jitteriness of the motions can result in higher diversity and higher beat consistency, as discussed in \cref{sec:quant}. Our network, generates smooth motions that help it score better in the facial metrics, where the jitteriness is penalized more. Additionally, the lower scores in FMD and FED can be explained by the dullness of the alternative methods' motions. This dullness, or staying relatively close to mean pose leads to the models on average performing well in the FED and FMD measures. This dull motion can be seen in the sample sequence shown in \cref{fig:ablat}.

\subsection{User Study}
The quantitative metrics can give a good idea of which method performs better and some like the Fr\'echet distances are intended to be consistent with human perception and judgment. However, these metrics still fall short of replacing humans and do not mirror human quality assessment perfectly. For this reason, we also conduct a user study of our method, the baseline methods, and our ablation studies. 

The study consisted of 10 participants who were asked to grade videos based on the quality of the generated co-speech gestures and facial motions. 
The SHOW dataset consists of 4 different speakers, for each speaker 3 audios of were randomly chosen to be used for comparison for a total of 12 different audios. We only chose audios from the longer videos which were about 10 seconds long.
We generate videos with each of the baseline methods, TalkSHOW \cite{yi2022generating}, DiffGesture \cite{zhu2023taming}, LS3DCG \cite{ls3dcg}, and we also generate videos for our ablation study methods, the combined and separated methods as described in \cref{sec:ablation}. The users were instructed to give grades of 1 to 5, with 5 being the best, to the following criteria: realism of the gestures, smoothness of gestures, synchronization between the speech audio and the gestures, synchronization between speech audio and the lip movements, and the realism of the facial expressions. 

The results of the user study, in \cref{tab:user}, shows our method performs better than all baselines and ablation study methods, only the ground truth performs better. Both combined and separated methods performed better than two of the baselines but worse than TalkSHOW, showcasing that the addition of the adapter modules significantly increase the performance of the base model enabling significant gains in realism of generated gestures and facial motions.

\section{Limitations and Future Work}
There are some limitations with the method that provide several future directions for the work. A major limitation of the work is the requirement for a seed gesture, this seed gesture not only sets the initial pose but can also have negative consequences on the entire sequence. If the initial pose is bad or very dissimilar to the training data, the resulting motion can be extreme or get stuck in unnatural poses, such as the head pointing downward or arms staying widely spread apart. This poor initialization can also lead to artifacts in the sequence if the initial pose or expression is bad, a problem that is not present in the LS3DCG, TalkSHOW, or DiffSHEG methods. To eliminate the need for a seed gesture for interpolating between clips, RePaint \cite{repaint} offers one possible solution. This approach was adopted by DiffSHEG \cite{chen2024diffsheg} and our architecture could be adapted to include this method. In addition to eliminating the seed gesture, expanding the dataset to include more speakers with a more diverse range of presentation could be used to help the network avoid getting stuck in poses that occur heavily in the training data.

Additionally, our quantitative results and user study showcase another limitation that has possibility for improvement: the design of metrics to measure human speech motion. The current design of metrics do not accurately mimic the response of human observers and over reliance on these metrics could result in methods that actually perform worse but test better. We believe that creating more sophisticated metrics to identify natural looking human speech motion is an exciting direction with a lot of potential.

\section{Conclusion}
In this work we have presented a method for utilizing adapter modules to jointly model two weakly correlated tasks. Our diffusion model is able to create state-of-the-art co-speech gestures jointly with an expressive talking face while maintaining a small parameter count in relation to baseline methods. Our experiments and user study have shown that our method can outperform existing methods to create realistic gestures and expressions with believable talking face from a driving speech audio. 

%%%%%%%%% REFERENCES
{\small
\bibliographystyle{ieee_fullname}
\bibliography{main}
}

\end{document}